\documentclass[letterpaper, 10 pt, journal, twoside]{IEEEtran}  % Comment this line out if you need a4paper

\IEEEoverridecommandlockouts                              % This command is only needed if 
\usepackage[%dvipdfm,
bookmarks=true,
bookmarksnumbered=true,
bookmarkstype=toc,%
pdfauthor={Liao Wu},%
pdftitle={THMS},%
pagebackref=false,
colorlinks=true,
citecolor=red,
linkcolor=blue,
anchorcolor=red]{hyperref}

\usepackage{graphicx} % for pdf, bitmapped graphics files
\usepackage{epsfig} % for postscript graphics files
\usepackage{mathptmx} % assumes new font selection scheme installed
\usepackage{times} % assumes new font selection scheme installed
\usepackage{amsmath} % assumes amsmath package installed
\usepackage{amssymb}  % assumes amsmath package installed
\usepackage{booktabs} %for table
\usepackage{multirow}
\usepackage{cite}
\usepackage{xcolor}
\usepackage{amsmath}
\usepackage{hyperref}
\DeclareMathOperator*{\argmax}{arg\,max}
\DeclareMathOperator*{\argmin}{arg\,min}
\newtheorem{remark}{Remark}

\title{\LARGE \bf
	Camera Frame Misalignment in a Teleoperated Eye-in-Hand Robot: Effects and a Simple Correction Method
}

\author{Liao~Wu,~\IEEEmembership{Member,~IEEE},~Fangwen~Yu,~\IEEEmembership{Member,~IEEE},~Thanh~Nho~Do,~\IEEEmembership{Member,~IEEE},~and~Jiaole~Wang% <-this % stops a space
	\thanks{Manuscript received...}
	\thanks{L. Wu is with the School of Mechanical and Manufacturing Engineering, University of New South Wales, Sydney, Australia. {\tt\small dr.liao.wu@ieee.org}}
	\thanks{F. Yu is with the Center for Brain Inspired Computing Research and the Department of Precision Instrument at Tsinghua University, Beijing, China. He was a visiting PhD student at Queensland University of Technology, Brisbane, Australia. \tt\small yufangwen@tsinghua.edu.cn}
	\thanks{T. N. Do is with the Graduate School of Biomedical Engineering and the Tyree Foundation Institute of Health Engineering (IHealthE), University of New South Wales, Sydney, Australia. {\tt\small tn.do@unsw.edu.au}}
	\thanks{J. Wang is with the School of Mechanical Engineering and Automation, Harbin Institute of Technology, Shenzhen 518055, China. \tt\small wangjiaole@hit.edu.cn}
	\thanks{Corresponding authors: L.~Wu {\tt\small dr.liao.wu@ieee.org} and J.~Wang {\tt\small wangjiaole@hit.edu.cn}}
}

\begin{document}

\maketitle
%\thispagestyle{empty}
%\pagestyle{empty}

%%%%%%%%%%%%%%%%%%%%%%%%%%%%%%%%%%%%%%%%%%%%%%%%%%%%%%%%%%%%%%%%%%%%%%%%%%%%%%%%
\begin{abstract}
Misalignment between the camera frame and the operator frame is commonly seen in a teleoperated system and usually degrades the operation performance.
The effects of such misalignment have not been fully investigated for \textit{eye-in-hand} systems - systems that have the camera (\textit{eye}) mounted to the end-effector (\textit{hand}) to gain compactness in confined spaces such as in endoscopic surgery.
This paper provides a systematic study on the effects of the camera frame misalignment in a teleoperated \textit{eye-in-hand} robot and proposes a simple correction method in the view display.
A simulation is designed to compare the effects of the misalignment under different conditions.
Users are asked to move a rigid body from its initial position to the specified target position via teleoperation, with different levels of misalignment simulated.
It is found that misalignment between the input motion and the output view is much more difficult to compensate by the operators when it is in the orthogonal direction ($\sim$40s) compared with the opposite direction ($\sim$20s).
An experiment on a real concentric tube robot with an \textit{eye-in-hand} configuration is also conducted.
Users are asked to telemanipulate the robot to complete a pick-and-place task.
Results show that with the correction enabled, there is a significant improvement in the operation performance in terms of completion time (mean 40.6\%, median 38.6\%), trajectory length (mean 34.3\%, median 28.1\%), difficulty (50.5\%), unsteadiness (49.4\%), and mental stress (60.9\%). 
A video demonstrating the experiments can be found at \href{https://youtu.be/EQILlwoS_Cw}{https://youtu.be/EQILlwoS\_Cw}.
\end{abstract}


%%%%%%%%%%%%%%%%%%%%%%%%%%%%%%%%%%%%%%%%%%%%%%%%%%%%%%%%%%%%%%%%%%%%%%%%%%%%%%%%
\section{Introduction}

\begin{figure}[tb]
	\centering
	%  \framebox{\parbox{3in}{We suggest that you use a text box to insert a graphic (which is ideally a 300 dpi TIFF or EPS file, with all fonts embedded) because, in an document, this method is somewhat more stable than directly inserting a picture.}}
	\includegraphics[width=60mm]{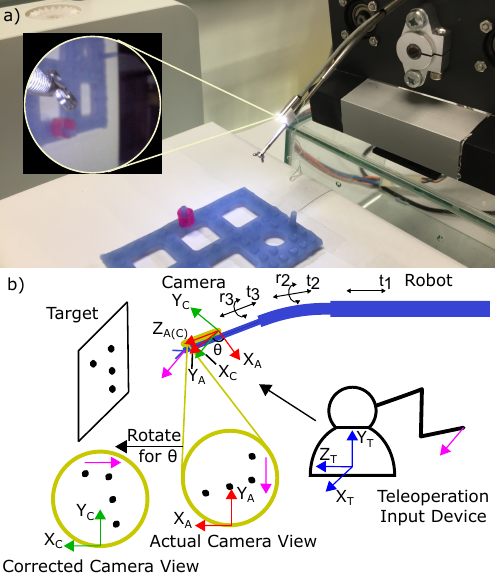}
	\caption{An example of camera frame misalignment in a teleoperated \textit{eye-in-hand} robot. a) A tubular robot with a camera mounted to the end-effector. The image returned by the camera is not consistent with the operator's view as the camera is rotated about its axis. b) When the operator applies a motion in the left direction (as indicated by the pink arrow on the operand), the camera also moves left (as indicated by the pink arrow on the camera), and the objects in the image should be viewed as if they move right (as indicated by the pink arrow in the corrected camera view) if no misalignment exists. However, the actual image moves downwards (as indicated by the pink arrow in the actual cameral view) due to misalignment of the actual camera frame \{$X_A$, $Y_A$, $Z_A$\} with respect to the teleoperator frame \{$X_T$, $Y_T$, $Z_T$\}.}
	\label{fig:schematic}
\end{figure}

Teleoperation is an effective way to control robots in complex environments as it integrates the superiority of human intelligence with the advantages of the mechatronic body of robots \cite{niemeyer2016telerobotics}.
Since the first development in the 1950s, teleoperated robotic systems have been widely used for space operations \cite{hirzinger1994rotex}, search and rescue \cite{murphy2004human}, rehabilitation and surgical applications \cite{guthart2000intuitive,razjigaev2017teleoperation,taylor1995telerobotic,razjigaev2021snakeraven,razjigaev2022end}, and others.

In a typical telerobotic system, visual display is used as the main, and sometimes the only, feedback for the human operator.
One or more cameras are usually attached to the slave system (the remote part that is teleoperated by the master device and directly interacts with the environment) to provide visual feedback.
Depending on the spatial relationship between the camera and the slave manipulator, the systems can be classified into two categories: \textit{eye-to-hand}, meaning that the camera is mounted independently of the end-effector of the manipulator and will not move with its motions \cite{yu2016development,guthart2000intuitive,kim2015effects,chintamani2009improved,taylor1995telerobotic,draelos2017teleoperating}, and \textit{eye-in-hand}, meaning that the camera is rigidly attached to the end-effector of the manipulator \cite{wu2015model,wu2016towards,rakita2018autonomous,wu2019safety}.
In general, the \textit{eye-to-hand} configuration is more widely used for teleoperated manipulators, because it provides view of the major part of the manipulator interacting with the environment and is more intuitive for the human operator who is also ``configured" as \textit{eye-to-hand}.
The advantages of the \textit{eye-in-hand} system, however, lie in its ability to provide more dynamic observation of the site where the interactions of the end-effector occur, and its being more compact for deployment in confined spaces such as in endoscopic surgery.

This work was motivated by a problem of camera frame misalignment encountered in a teleoperated \textit{eye-in-hand} tubular robot developed for transnasal procedures \cite{wu2016towards}, as illustrated in Fig. \ref{fig:schematic} where a camera is rigidly attached to the end-effector of a tubular robot. 
An operator uses an input device to telemanipulate the robot while observing the visual display captured by the camera. 
The motion of the end-effector is linearly mapped by the motion of the input device via a differential kinematic model \cite{wu2017development}.

The robot consists of three concentric tubes, with the outermost and the innermost tubes being straight and the middle tube pre-curved.
A gripper and a camera are rigidly mounted to the innermost tube.
Each tube can be rotated and translated independently through transmission mechanisms controlled by motors.
As the rotation of the outermost tube does not contribute to the end-effector's motion, the total number of degrees of freedom is five, i.e., $t_1$, $t_2$, $r_2$, $t_3$, and $r_3$ in Fig. \ref{fig:schematic}.

There are three main characteristics in the kinematics of the robot \cite{burgner2014telerobotic,dupont2010design} that are relevant to the misalignment studied in this paper:
\begin{itemize}
	\item As the camera is rigidly attached to the innermost tube, the orientation of the camera may change dynamically when the robot is teleoperated by an input device. 
	\item The change of orientation of the camera could cause a significant rotation about its axis, because the innermost tube can be rotated about its axis independently. Moreover, as the innermost tube is made of spring, there is torsional compliance when the rotation is transmitted from one end (motor) to another (camera), adding uncertainty to the rotation angle.
	\item The change of orientation of the camera could also involve variation in the tangential direction of the axis, as determined by the rotation of the middle tube that is pre-curved. Note that, however, the variation is bounded to a certain range ($\pm 45^\circ$) as the length of the curve in the middle tube is fixed.
\end{itemize}

%As the camera is rigidly attached to the end-effector, the orientation of the camera may change dynamically while the end-effector is in motion, particularly when there is rotation about the axis of the end-effector.
%There may also be misalignment between the camera frame and the teleoperation frame in the initial state.
Due to the misalignment between the camera frame and the teleoperation frame, the motion observed in the actual camera view can be in a significantly different direction from the input motion of the operator.
Taking the case shown in Fig. \ref{fig:schematic} for instance, when the input device is moved in the positive direction of $\mathbf{X}_T$, the image in the actual camera view shows a motion downwards due to the misalignment between the actual camera frame $\{A\}$ and the teleoperation frame $\{T\}$.
An initial calibration may align the camera frame with the teleoperation frame at the start state.
However, as the camera's orientation may change dynamically with the motion of the robot and the uncertainty brought by the compliance of the innermost tube, misalignment may still appear along the process of teleoperation.
Therefore, a one-off calibration may not be sufficient.

Studies have shown that misalignment in teleoperation may significantly affect the performance of the operator\cite{ames2006evaluation,hiatt2006coordinate,murphy2004human,chen2007human}.
%koreeda2015development,kim2015effects,draelos2017teleoperating,ellis2012human 
A correction can be applied to mitigate the effect.
One way to correct the misalignment is to transform the motion of the master to the camera frame first and then map the movement of the end-effector in the motion mapping step.
While this method seems feasible, it has two drawbacks: 1) it does not apply to systems that use kinematically similar mechanisms and do not allow such transformation in motion mapping \cite{niemeyer2016telerobotics}, e.g., in a mechanically linked teleoperated system; 2) the \textit{a priori} knowledge about the interaction site established in the operator's teleoperation frame may be violated by the rotation of the camera view, requiring the operator to do mental rotation for the task, which may increase the burden of the operator and the time to complete the task \cite{shepard1971mental}. 

Therefore, this paper proposes an alternative approach that corrects the misalignment in the camera view display (Fig.~\ref{fig:schematic}.b), leaving the motion mapping unchanged and keeping the \textit{a priori} knowledge about the teleoperated environment for the first time.
Moreover, the effects of camera frame misalignment in this configuration are systematically studied, and the effectiveness of the proposed method is validated by carefully designed user studies.

\subsection{Related Work}
\subsubsection{Effects of the misalignment}
Ames et al. \cite{ames2006evaluation} evaluated the performance of surgeons with altering laparoscope positions with respect to the working instruments.
Although the operation was not teleoperated, the setup shared the same characteristics with a teleoperated \textit{eye-to-hand}  system. 
Configurations with the angle between the laparoscope and the baseline position being $0^\circ$, $45^\circ$, $90^\circ$, $135^\circ$, and $180^\circ$ were tested.
Results showed that the performance of the participants deteriorated with the increase of deviation of the laparoscope angle from baseline for both novice and experienced surgeons.

Ellis et al. \cite{ellis2012human} studied the effects of misalignment between display and control axes for pitch, roll, and yaw rotations in a vitual reality system.
By inviting 20 subjects to complete a three-phase task, they found a peak drop of performance around $120^\circ$ for all the three axes.
They also found that the effects of the control-display misalignment was anisotropic with roll being more significantly susceptible. 

Kim et al. \cite{kim2015effects} used the Raven system and the da Vinci Research Kit (dVRK), two typical teleoperated \textit{eye-to-hand} robots, to evaluate the effects of the misalignment between the master input and the slave tool motion.
They found that operators were able to compensate for up to approximately $20^\circ$ in tool orientation and camera viewpoint misalignment.

All the above works studied the misalignment problem in the \textit{eye-to-hand} configuration, while few of them have investigated the \textit{eye-in-hand} system.
An exception is the research by Macedo et al. \cite{macedo1998effect} which evaluated the effects of misalignment between a 2D display and a 2D joystick with respect to different rotation angles.
Through experiments on 18 participants, they found that a simple compensation worked for all the rotation angles, whereas when there was no compensation, two peak values appeared at $90^\circ$ and $270^\circ$ without visual/display compatibility.
Although the work in \cite{macedo1998effect} is relevant to the present study, it is limited to a 2D system and a more comprehensive investigation of the effects is needed.

\subsubsection{Correction methods for the misalignment}
In general, two types of correction have been proposed for the misalignment, one in the motion mapping step and one in the view display.
Draelos et al. \cite{draelos2017teleoperating} proposed a framework called Arbitrary Viewpoint Robotic Manipulation (AVRM) with which the misalignment in the tool motions can be corrected in an arbitrary viewpoint.
More correction methods are in improving the view display.
For example, Dejone et al. \cite{dejong2004improving,dejong2011mental} proposed to reduce the teleoperation mental workload by optimizing component arrangement in an interface.
Augmented reality cues have also been suggested to assist the operator's telemanipulation \cite{cao2000augmented,chintamani2009improved}.
Other methods include deforming the 2D image of an endoscope to generate a pseudo-viewpoint \cite{koreeda2015development} and having a second camera-in-hand robot arm that provides dynamic moving viewpoint \cite{rakita2018autonomous}.
However, all of these methods mentioned have only dealt with the misalignment problem in a teleoperated \textit{eye-to-hand} system. 
\subsection{Contribution}
Based on the introduction above, it can be seen that the effects of the camera frame misalignment have mainly been studied for \textit{eye-to-hand} systems.
For \textit{eye-in-hand} configurations, study has been limited to 2D cases only.
Moreover, no correction methods that apply to the \textit{eye-in-hand} configurations have been proposed.
Therefore, the main contribution of this paper is twofold:
\begin{itemize}
	\item A simple correction method in the view display is proposed and validated through comprehensive user studies;
	\item The effects of the camera frame misalignment in a teleoperated \textit{eye-in-hand} robot are investigated systematically for the first time.
\end{itemize}

\section{A Simple Correction Method for the Camera Frame Misalignment} \label{sec:correction}

\begin{figure}[tb]
	\centering
	\includegraphics[width=30mm]{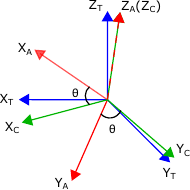}
	\caption{A simple correction method for the camera frame misalignment. Frames \{T\}, \{A\}, and \{C\} represent the teleoperator's, the actual, and the corrected frame, respectively. The corrected frame is obtained by rotating the actual frame about its $\mathbf{Z}$ axis for an angle of $\theta$.}
	\label{fig:method}
\end{figure}

Suppose the frames associated with the teleoperation input device and the actual camera are \{$\mathbf{X}_T$, $\mathbf{Y}_T$, $\mathbf{Z}_T$\} and \{$\mathbf{X}_A$, $\mathbf{Y}_A$, $\mathbf{Z}_A$\}, respectively, as shown in Fig. \ref{fig:method}.
The corrected camera frame is denoted as \{$\mathbf{X}_C$, $\mathbf{Y}_C$, $\mathbf{Z}_C$\}.
We propose to correct the camera view by rotating it about its $\mathbf{Z}$ axis; therefore $\mathbf{Z}_A$ and $\mathbf{Z}_C$ are identical. 
Note that this rotation will not resolve the misalignment between the $\mathbf{Z}$ axes of the operation frame and the camera frame, which can exist when the camera's tangential direction deviates from the operator's $\mathbf{Z}$ axis. 
However, rotating the camera image only, due to its resemblance to the original view, will require the least learning effort for the users to adapt from the usual display to the new display.
Moreover, in the robot motivating this study, the misalignment in the $\mathbf{Z}$ direction is bounded to a certain range ($\pm 45^\circ$).
We hypothesize that this simple rotation should be effective for the studied configurations.
The angles between $\mathbf{X}_A$($\mathbf{Y}_A$) and $\mathbf{X}_C$($\mathbf{Y}_C$) is denoted as $\theta$.

The aim of the correction is to ensure the image displayed in the corrected camera frame is as close as the one observed from the teleoperator's view. Therefore, we define the correction of the misaligned camera frame as a problem of minimizing the ``distances" between $\mathbf{X}_C$ and $\mathbf{X}_T$ as well as between $\mathbf{Y}_C$ and $\mathbf{Y}_T$. Different metrics can be used to evaluate the ``distance" between two vectors; in this problem, as the angle between the two coordinate axes is the focus, we use the cosine distance $d(\cdot,\cdot)$, which is defined as
\begin{equation}\label{eq:cosine}
	d(\mathbf{a},\mathbf{b})=1-s(\mathbf{a},\mathbf{b})=1-\cos(\phi)=1-\frac{\mathbf{a}\cdot\mathbf{b}}{||\mathbf{a}|| ||\mathbf{b}||}
\end{equation}
where $\mathbf{a}$ and $\mathbf{b}$ are two vectors, $s(\cdot,\cdot)$ is the cosine similarity, and $\phi$ is the angle between $\mathbf{a}$ and $\mathbf{b}$.

Given the directions of \{$\mathbf{X}_T$, $\mathbf{Y}_T$\} and \{$\mathbf{X}_A$, $\mathbf{Y}_A$\}, the correction for the camera frame misalignment can be defined as finding an angle $\theta$ that satisfies 
\begin{equation} \label{eq:argmin}
	\underset{\theta}{\argmin}{\quad d(\mathbf{X}_T,\mathbf{X}_C)+d(\mathbf{Y}_T,\mathbf{Y}_C)}.
\end{equation}
According to (\ref{eq:cosine}), this is equal to finding $\theta$ that satisfies
\begin{equation} \label{eq:argmax}
	\underset{\theta}{\argmax}{\quad s(\mathbf{X}_T,\mathbf{X}_C)+s(\mathbf{Y}_T,\mathbf{Y}_C)}.
\end{equation}
Since the axis vectors are all unit vectors, (\ref{eq:argmax}) can be simplified as
\begin{equation} \label{eq:argmax2}
	\underset{\theta}{\argmax}{\quad \mathbf{X}_T^T\mathbf{X}_C+\mathbf{Y}_T^T\mathbf{Y}_C}.
\end{equation}

As $\mathbf{X}_C$ and $\mathbf{Y}_C$ are rotated from $\mathbf{X}_A$ and $\mathbf{Y}_A$ about $\mathbf{Z}_A$ for $\theta$, respectively, we have
\begin{equation}
	\left[\begin{matrix}
	\mathbf{X}_C\\
	\mathbf{Y}_C
	\end{matrix}\right]=\left[\begin{matrix}
	\cos(\theta) & \sin(\theta)\\
	-\sin(\theta) & \cos(\theta)
	\end{matrix}\right]\left[\begin{matrix}
	\mathbf{X}_A\\
	\mathbf{Y}_A
	\end{matrix}\right],
\end{equation}
i.e.,
\begin{eqnarray}
	\mathbf{X}_C=\mathbf{X}_A\cos(\theta)+\mathbf{Y}_A\sin(\theta)\label{eq:XC}\\
	\mathbf{Y}_C=\mathbf{Y}_A\cos(\theta)-\mathbf{X}_A\sin(\theta)\label{eq:YC}.
\end{eqnarray}
Substituting (\ref{eq:XC}) and (\ref{eq:YC}) into (\ref{eq:argmax2}), we have
\begin{equation}\label{eq:argmax3}
	\underset{\theta}{\argmax}{\quad (\mathbf{X}_T^T\mathbf{Y}_A-\mathbf{Y}_T^T\mathbf{X}_A)\sin(\theta)+(\mathbf{X}_T^T\mathbf{X}_A+\mathbf{Y}_T^T\mathbf{Y}_A)\cos(\theta)}.
\end{equation}
It can be verified that the solution to (\ref{eq:argmax3}) is given by
\begin{equation}\label{eq:solution}
	\theta = \text{atan2}(\mathbf{X}_T^T\mathbf{Y}_A-\mathbf{Y}_T^T\mathbf{X}_A, \mathbf{X}_T^T\mathbf{X}_A+\mathbf{Y}_T^T\mathbf{Y}_A)
\end{equation}
where the order $\phi=\text{atan2}(\sin(\phi),\cos(\phi))$ is used.
This means that given \{$\mathbf{X}_T$, $\mathbf{Y}_T$\} and \{$\mathbf{X}_A$, $\mathbf{Y}_A$\}, we can use (\ref{eq:solution}) to calculate the rotation angle needed for the correction.

\begin{remark}\label{rm:forward}
	If the actual camera frame is in minimal misalignment with the teleoperation input device frame, then the angle between $\mathbf{X}_T$ and $\mathbf{Y}_A$ should be the same as the one between $\mathbf{Y}_T$ and $\mathbf{X}_A$.
\end{remark}

This can be proved by considering the correction angle derived above. 
If the actual camera frame is in minimal misalignment with the teleoperation input device frame, then the correction angle needed should be zero.
According to (\ref{eq:solution}), this requires $\mathbf{X}_T^T\mathbf{Y}_A=\mathbf{Y}_T^T\mathbf{X}_A$, i.e., the angle between $\mathbf{X}_T$ and $\mathbf{Y}_A$ is the same as the one between $\mathbf{Y}_T$ and $\mathbf{X}_A$.

\begin{remark}\label{rm:example}
	The angle between $\mathbf{X}_T$ and $\mathbf{Y}_A$ equaling the one between $\mathbf{Y}_T$ and $\mathbf{X}_A$ does not guarantee that the camera frame is in minimal misalignment.
\end{remark}

This is because there is a possibility that $\mathbf{X}_T^T\mathbf{X}_T+\mathbf{Y}_T^T\mathbf{Y}_A<0$, which results in $\theta$ calculated by (\ref{eq:solution}) to be $\pi$ instead of zero.
One such case is when $\mathbf{X}_A=-\mathbf{X}_T$ and $\mathbf{Y}_A=-\mathbf{Y}_T$.
We then have
\begin{equation}
\theta = \text{atan2}(\underset{=0}{\underline{-\mathbf{X}_A^T\mathbf{Y}_A+\mathbf{Y}_T^T\mathbf{X}_T}}, \underset{<0}{\underline{-\mathbf{X}_T^T\mathbf{X}_T-\mathbf{Y}_T^T\mathbf{Y}_T}})=\pi.
\end{equation}

\begin{remark}\label{rm:inverse}
	The camera frame is in minimal misalignment if $\mathbf{X}_A$ aligns with $\mathbf{X}_T$ or $\mathbf{Y}_A$ aligns with $\mathbf{Y}_T$.
\end{remark}

This can be verified by evaluating (\ref{eq:solution}). When $\mathbf{X}_A=\mathbf{X}_T$, (\ref{eq:solution}) becomes
\begin{equation}
	\theta = \text{atan2}(\underset{=0}{\underline{\mathbf{X}_A^T\mathbf{Y}_A-\mathbf{Y}_T^T\mathbf{X}_T}}, \underset{\ge 0}{\underline{\mathbf{X}_T^T\mathbf{X}_T+\mathbf{Y}_T^T\mathbf{Y}_A}})=0.
\end{equation}
When $\mathbf{Y}_A=\mathbf{Y}_T$, (\ref{eq:solution}) becomes
\begin{equation}
\theta = \text{atan2}(\underset{=0}{\underline{\mathbf{X}_T^T\mathbf{Y}_T-\mathbf{Y}_A^T\mathbf{X}_A}}, \underset{\ge 0}{\underline{\mathbf{X}_T^T\mathbf{X}_A+\mathbf{Y}_T^T\mathbf{Y}_T}})=0.
\end{equation}
Therefore, the camera frame is in minimal misalignment in these two cases. 
This indicates that the camera frame does not need to be corrected if one of the camera view axis aligns with the corresponding axis of the operator's frame.

Note that (\ref{eq:solution}) assumes the $\mathbf{X}$ and $\mathbf{Y}$ axes of the camera are known during the operation.
This can be achieved by using measurement devices such as a tracking system or algorithms such as camera pose estimation.
Eq. (\ref{eq:solution}) should be applied constantly when the camera's orientation changes dynamically during the operation.
If the camera's orientation is fixed, the method can also be used as calibration at the beginning of the process.

%\section{Evaluation of the Effect of Camera Frame Misalignment and Validation of the Correction Method}
%Two groups of experiments, one in a simulated environment and one on a real robotic system, were tested by a group of participants to evaluate the effect of camera frame misalignment and validate the proposed correction method. 
%All the participants provided informed consent. 
%This protocol was approved by the Australian National Health and Medical Research Council (NHMRC) Registered Committee Number EC00171 under Approval Number 1800000891.

\section{Evaluation of the Effects of Camera Frame Misalignment in a Simulated Environment} \label{sec:simulation}
In this section, we investigate the effects of camera frame misalignment in a teleoperated \textit{eye-in-hand} robot by inviting participants to conduct a series of experiments, both with and without camera correction as introduced in Sec. \ref{sec:correction}, under different conditions in a simulated environment.
A three-way mixed experiment was designed for this evaluation and the details are described as follows.

\subsection{Participants}
A total of 32 participants (20 males and 12 females) volunteered for this experiment. 
The ages of the participants range between 22 and 36 (M = 29.78, SD = 3.80). 
The occupations include 30 engineering researchers/students and two business students. 
All of the participants are right-handed, and have normal visual acuity and no dyschromatopsia (color blindness).
None but one participant had experience with the 3D joystick used in this experiment (the exception only had limited experience in using the 3D joystick to remote-control helicopters).
All the participants provided informed consent. 
This protocol was approved by the Australian National Health and Medical Research Council (NHMRC) Registered Committee Number EC00171 under Approval Number 1800000891.

\subsection{Methods}
\begin{figure*}[tb]
	\centering
	\includegraphics[width=120mm]{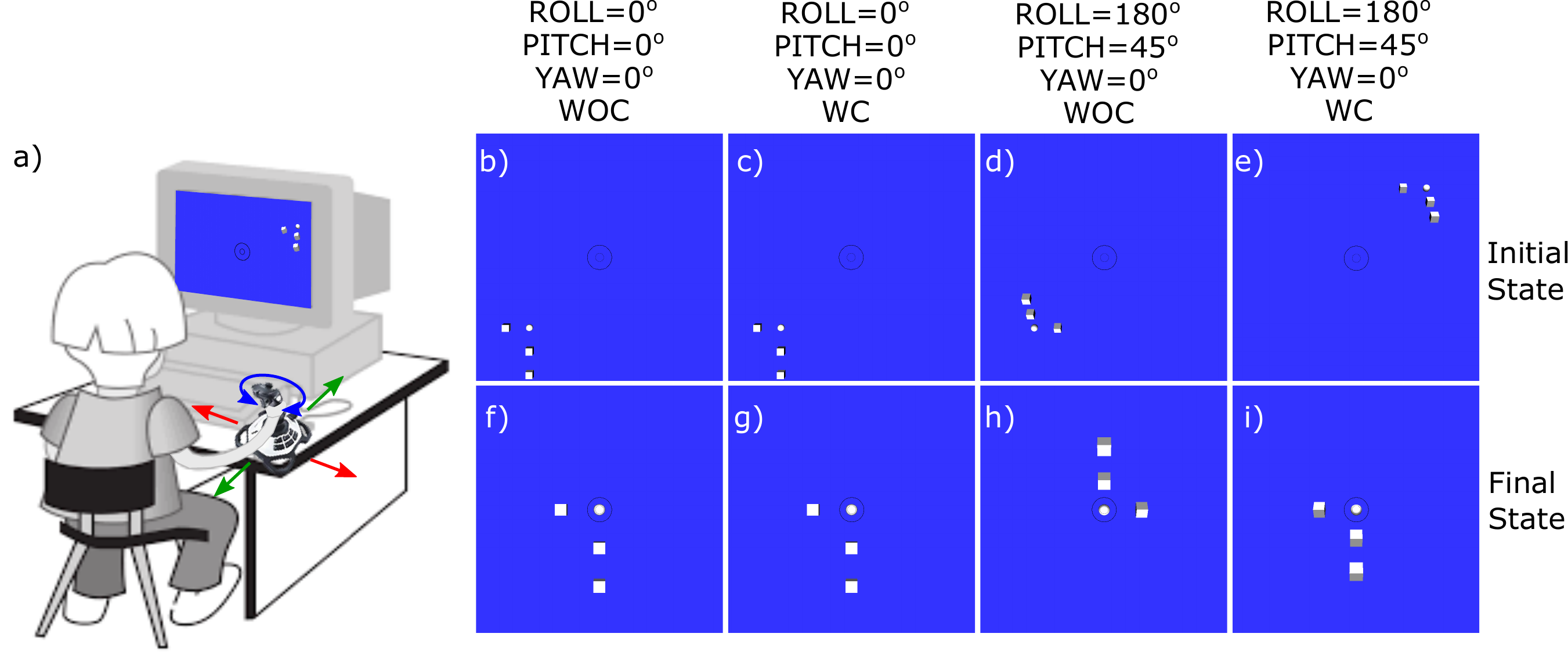}
	\caption{Teleoperation experiments in a simulated environment. a) Experimental setup. The operator uses a 3D joystick to manipulate a quadruplet constituted by a sphere and three cubes in a MATLAB simulation. The task is to move the camera from the initial position to a position where the 2D projection of the sphere lies between the two concentric circles at the center of the camera view. b) - i) Example images presented to the operator under different conditions. Note that the states between WOC and WC when $ROLL=0^\circ,PITCH=0^\circ,YAW=0^\circ$ are the same as there is no misalignment under this condition.}
	\label{fig:simulation}
\end{figure*}

The participant was asked to telemanipulate a quadruplet in a simulated environment by maneuvering a 3D joystick (Extreme 3D Pro, Logitech), as illustrated in Fig. \ref{fig:simulation} (a). 
A blue square background was presented to the participant simulating the view of a camera that was sitting on the end-effector of a robot.
A sphere and three cubes, forming a quadruplet with fixed position in the 3D space, were observed by the camera.
The cubes were arranged in a way helping identify the orientation of the quadruplet. 
Examples of the initial position of the quadruplet in the camera view are illustrated in Fig. \ref{fig:simulation} (b) - (e).
The task was to move the camera from the initial position to a position where the 2D projection of the sphere lies between the two concentric circles at the center of the camera view, as seen in Fig. \ref{fig:simulation} (f) - (i).

\begin{figure*}[tb]
	\centering
	\includegraphics[width=120mm]{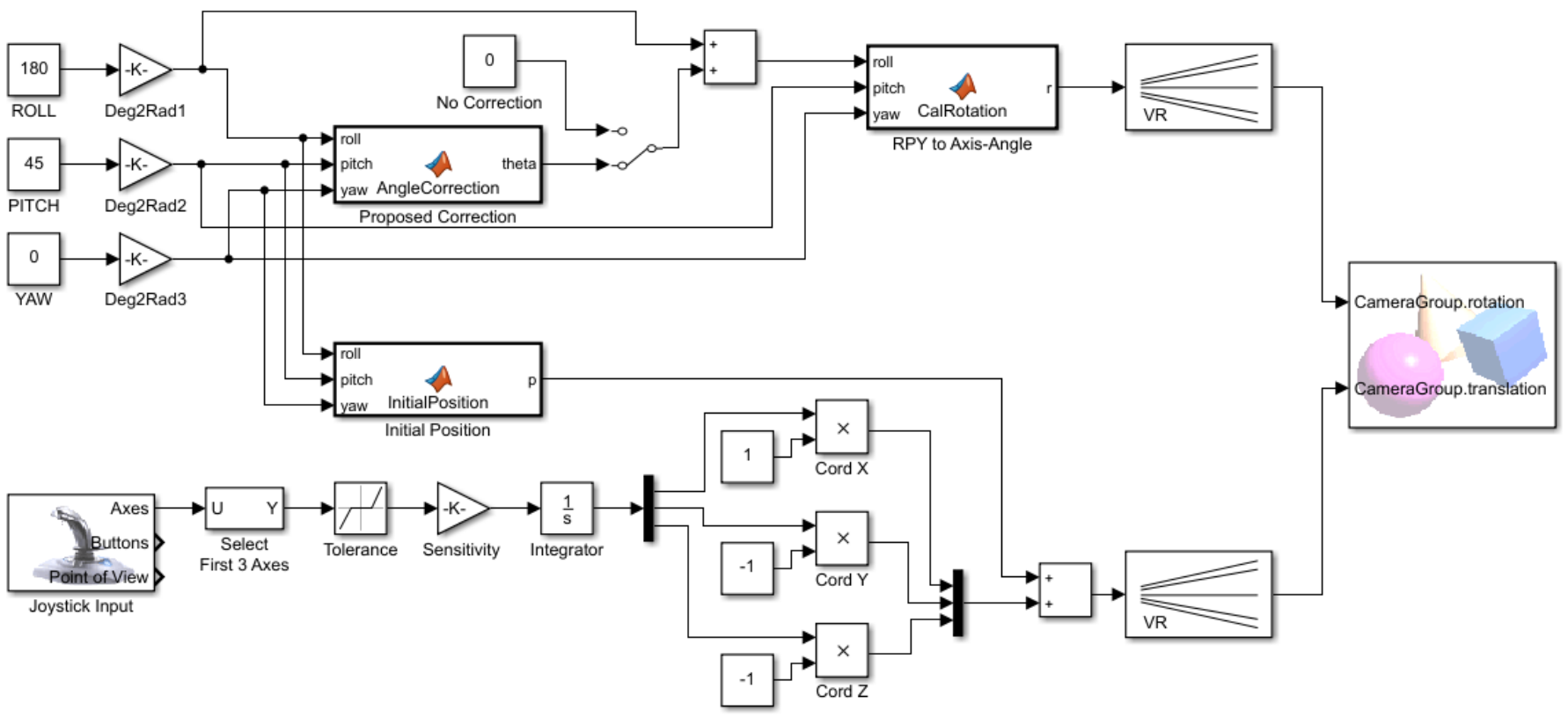}
	\caption{Block diagram of the simulation algorithm. The orientation of the camera is determined by ROLL, PITCH, and YAW angles. Two conditions, WOC and WC, can be selected through a switch by the experimenter. The simulation runs in Simulink (9.0) 3D Animation on a Windows 10 platform with Intel Core i5-7300U CPU @ 2.60GHz 2.71GHz and 8.00GB RAM.}
	\label{fig:matlab}
\end{figure*}

The environment was built based on MATLAB 2017b and Simulink (9.0) 3D Animation provided by MathWorks, Inc.
The source code of the simulation can be found at \href{https://github.com/drliaowu/CameraMisalignmentCorrection}{https://github.com/drliaowu/CameraMisalignmentCorrection}.
The block diagram of the simulation is shown in Fig. \ref{fig:matlab}.
The orientation of the camera was determined by four parameters: \textit{roll (ROLL)}, \textit{pitch (PITCH)}, \textit{yaw (YAW)}, and \textit{with correction (WC) / without correction (WOC)}.
Under WOC condition, a rotation $r$ was generated by $Rot(y,YAW)Rot(x,PITCH)Rot(z,ROLL)$ and converted to the \textit{Axis-Angle representation} \cite{axisangle} before sent to the 3D Animation as the orientation of the camera.
Here $Rot(a,B)$ means a rotation about axis $a$ for angle $B$.
If the switch was set to WC condition, a correction angle $\theta$ was calculated using the algorithm proposed in Sec. \ref{sec:correction} and added to the roll angle before generating the rotation.
Specifically in the \textit{Proposed Correction} block in Fig. \ref{fig:matlab}, we first converted the RPY angles to a rotation matrix to extract the $\mathbf{X}$ and $\mathbf{Y}$ axes, and then implemented (\ref{eq:solution}) to obtain the correction angle.

\begin{figure*}[tb]
	\centering
	\includegraphics[width=100mm]{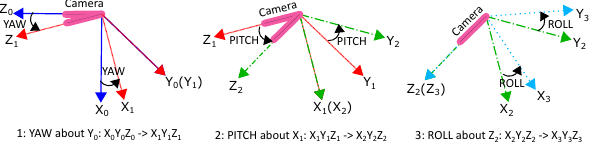}
	\caption{The RPY representation used in this paper. The three rotations transform the original frame $X_0Y_0Z_0$ to frame $X_3Y_3Z_3$ following the shown order. As the $\mathbf{Z}$ axis is located along the camera, it can be seen that the YAW and PITCH angles determine the direction of the axis of the camera, while the ROLL angle describes the rotation of the camera about its own axis.}
	\label{fig:rpy}
\end{figure*}

The RPY representation used in this paper is illustrated in Fig. \ref{fig:rpy}.
Under the specified order, the direction of the camera ($\mathbf{Z}$ axis) is only determined by the YAW and PITCH angles, while the ROLL angle describes the rotation of the camera about its own axis.
This decoupling brings convenience to the analysis of the effects of the misalignment, as it allows us to compare misalignment in the axial rotation with misalignment in the tangential direction of the axis. 

The three output signals of the joystick were sampled and integrated as the control commands for the $\mathbf{X}$, $\mathbf{Y}$, and $\mathbf{Z}$ motions of the camera.
Added to these signals was an initial position of the quadruplet relative to the camera.
The initial position was calculated based on the orientation of the camera so that the quadruplet was always seen in the camera view and the distance from the initial position to the final position was the same under different RPY angles.
Letting $\mathbf{R}$ be the rotation matrix of a given set of RPY angles, we calculated the initial position by
\begin{equation} 
	\mathbf{p}_{i} = \mathbf{R}\mathbf{v}_{o}+\mathbf{p}_{q}
\end{equation} 
where $\mathbf{v}_{o}=[30,30,140]^T$ is a fixed offset vector chosen empirically to maintain the quadruplet in the camera view and $\mathbf{p}_{q}=[0,0,-40]^T$ is the default position of the quadruplet with respect to the camera.
This was implemented in the \textit{Initial Position} block in Fig. \ref{fig:matlab}.

Note that the choice between WOC and WC does not affect the signal fed to the translation of the camera, but only impacts the rotation.
As a consequence, the same sequence of operations would result in exactly the same trajectories of the camera under the two conditions, although the observed trajectories would seem significantly different.
This feature is important for the fair comparison of the performance under the two conditions.
If there is little effect of the misalignment of the camera, similar completion times under the two conditions can be expected.

The 32 participants were randomly and equally divided into four groups, under the conditions ($PITCH=0^\circ,YAW=0^\circ$), ($PITCH=0^\circ,YAW=45^\circ$), ($PITCH=45^\circ,YAW=0^\circ$), and ($PITCH=45^\circ,YAW=45^\circ$), respectively.
In each group, the participant completed 48 trials with $ROLL$ being $0^\circ$, $45^\circ$, $90^\circ$, $135^\circ$, $180^\circ$, $225^\circ$, $270^\circ$, and $315^\circ$, and the switch connecting to WOC and WC (three repetitions under each combination of conditions, thus the total number of trials was $3\times8\times2=48$).  
To balance the carryover effects like learning and fatigue that are possible in within-subject experiments \cite{yoshimura1996historical}, Latin square was used to design the experiment.
The detailed sequence of conditions can be found in the raw data attached to this paper.
The simulation was run on a Windows 10 platform with Intel Core i5-7300U CPU @ 2.60GHz 2.71GHz and 8.00GB RAM.
Time of completion was recorded for each trial of task.

\subsection{Results}

A three-way repeated measures ANOVA was performed, with the three factors being PITCH-YAW (4 levels), ROLL (8 levels), and CORRECTION (2 levels).
Among them, ROLL and CORRECTION are two within-subject factors and PITCH-YAW is a between-subject factor. 
As the main interest of this study was on the effects of the correction method, the main effect of CORRECTION and the interaction effect between CORRECTION and the other factors were examined.

\subsubsection{Main effect of CORRECTION}
\begin{figure}[tb]
	\centering
	\includegraphics[width=40mm]{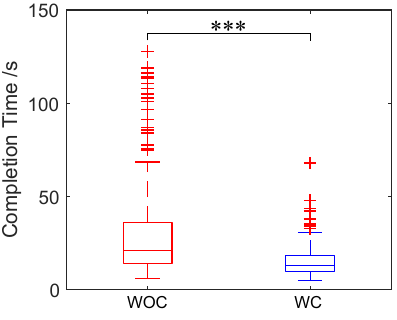}
	\caption{Completion time of the tasks WOC and WC. Each condition contains 256 data points. ***$p<0.001$.}
	\label{fig:simboxcor}
\end{figure}

The result of the main effect of CORRECTION is plotted in Fig. \ref{fig:simboxcor}.
The analysis compared the two levels of CORRECTION while not distinguishing the other factors.
There was a significant effect of CORRECTION on the completion time of the tasks, $F(1, 28)= 52.435$, $p<0.001$. 
A post hoc paired-samples \textit{t}-test revealed that with the correction algorithm enabled, the operators were able to complete the tasks much faster than without the correction algorithm ($15.16 \pm 1.05$ $s$ vs $29.67 \pm 2.72$ $s$, $t(255)=10.418$, $p<0.001$ ).

%\begin{figure}[tb]
%	\centering
%	\includegraphics[width=88mm]{Figures/simpolar}
%	\caption{Comparison.}
%	\label{fig:simpolar}
%\end{figure}

\subsubsection{Two-way interaction effect between CORRECTION and ROLL}
\begin{figure}[tb]
	\centering
	\includegraphics[width=60mm]{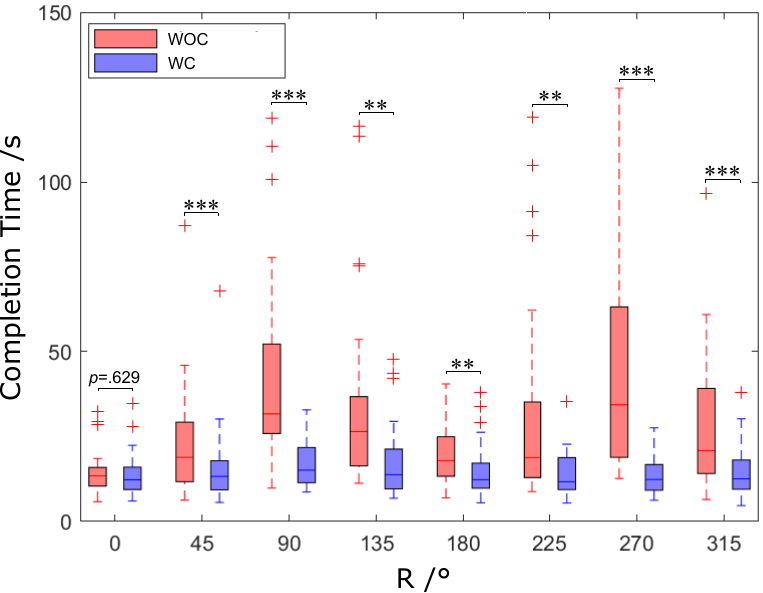}
	\caption{Completion time of the tasks WOC and WC under different ROLL angles. Each condition contains 32 data points. **$p<0.01$ and ***$p<0.001$.}
	\label{fig:simboxtotal}
\end{figure}

The result of the interaction effect between CORRECTION and ROLL is plotted in Fig. \ref{fig:simboxtotal}.
The analysis did not differentiate the levels of PITCH-YAW.
There was a significant interaction effect (with a Greenhouse-Geisser correction as the Mauchly's Test of Sphericity was significant) between CORRECTION and ROLL on the completion time of the tasks, $F(3.854,107.925)=11.681$, $p<0.001$.

Eight paired-samples \textit{t}-tests were used to make post hoc comparisons. 
As indicated in Fig. \ref{fig:simboxtotal}, there was a significant difference between WC and WOC in the conditions of nonzero ROLL angles, while the completion time was not significantly different when the ROLL angle was $0^\circ$.

\begin{figure}[tb]
	\centering
	\includegraphics[width=60mm]{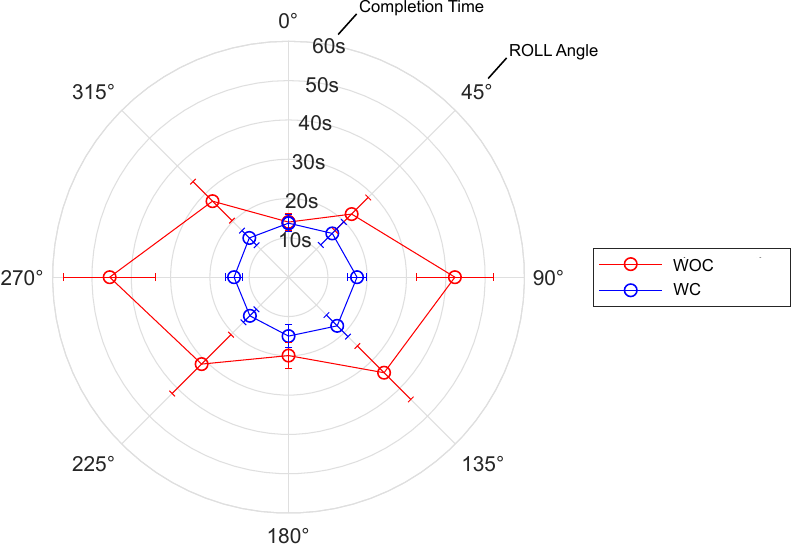}
	\caption{The interaction effect between CORRECTION and ROLL. The red line depicts the average completion time with respect to different ROLL angles without CORRECTION, while the blue line shows the results with CORRECTION. The error bars indicate 95\% confidence intervals.}
	\label{fig:simpolartotal}
\end{figure}

The pattern of the interaction effect between CORRECTION and ROLL can be seen more clearly in Fig. \ref{fig:simpolartotal}.
The red line depicts the average completion time with respect to different ROLL angles under WOC, while the blue line shows the results under WC. 
The error bars indicate 95\% confidence intervals.
A few observations can be made:
\begin{itemize}
	\item When the CORRECTION was enabled, the average completion time was similar (Min 13.74~s, Max 17.50~s) for all the eight ROLL angles.
	\item When the CORRECTION was disabled, however, the average completion time evinced significant disparities with different ROLL angles:
	\begin{itemize}
		\item When the ROLL angle was $0^\circ$, the completion time was the shortest (14.01~s) and similar as the one under WC (13.74~s).
		\item When the ROLL angle was $180^\circ$, the completion time increased slightly (19.94~s) and deviated from the one under WC (14.98~s).
		\item The completion time increased further with the ROLL angle being $45^\circ$ (22.70~s), $135^\circ$ (34.36~s), $225^\circ$ (19.94~s), and $315^\circ$ (27.34~s).
		\item The completion time reached the maximum when the ROLL angle was $90^\circ$ (42.31~s) and $270^\circ$ (45.48~s).
	\end{itemize}
\end{itemize}

\subsubsection{Two-way interaction effect between CORRECTION and PITCH-YAW}

There was no significant interaction effect between CORRECTION and PITCH-YAW, $F(3, 28) = 0.082$, $p = 0.969$.

\subsubsection{Three-way interaction effect of CORRECTION, ROLL, and PITCH-YAW}
\begin{figure}[tb]
	\centering
	\includegraphics[width=60mm]{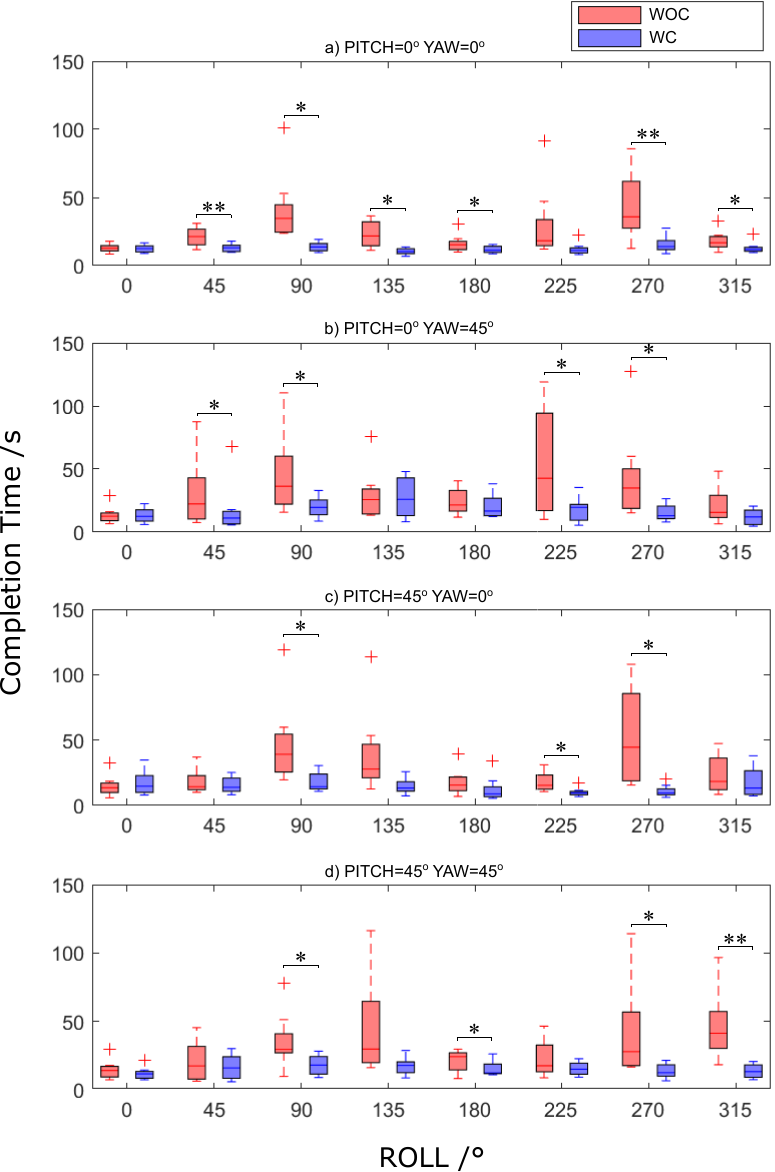}
	\caption{Completion time of the tasks with and without CORRECTION under different ROLL and PITCH-YAW angles. Each condition contains eight data points.  *$p<0.05$ and **$p<0.01$.}
	\label{fig:simbox}
\end{figure}

The result of the interaction effect of CORRECTION, ROLL, and PITCH-YAW is plotted in Fig. \ref{fig:simbox}.
There was a significant interaction effect (with a Greenhouse-Geisser correction as the Mauchly's Test of Sphericity was significant) between the three factors on the completion time of the tasks, $F(11.563,107.925)=2.272$, $p=0.014$.
The results of post hoc paired-samples \textit{t}-tests are also shown in Fig. \ref{fig:simbox}.
The following observations are obtained:
\begin{itemize}
	\item When the ROLL angle was $0^\circ$, the average completion time under WC and WOC showed no significant difference in all the four conditions of PITCH-YAW angles.
	\item When the ROLL angle was $45^\circ$, $135^\circ$, $180^\circ$, $225^\circ$, and $315^\circ$, the average completion time under WC and WOC was significantly different in some conditions of PITCH-YAW angles, while not significantly different in other conditions.
	\item When the ROLL angle was $90^\circ$ and $270^\circ$, the average completion time under WC and WOC was significantly different in all the four conditions of PITCH-YAW angles.
%	\item When the CORRECTION was enabled, the average completion time was similar (Min 13.74~s, Max 17.50~s) for all the eight ROLL angles.
%	\item When the CORRECTION was disabled, however, the average completion time evinced~significant disparities with different ROLL angles:
%	\begin{itemize}
%		\item When the ROLL angle was $0^\circ$, the completion time was the shortest (14.01~s) and similar as the one with CORRECTION (13.74~s).
%		\item When the ROLL angle was $180^\circ$, the completion time increased slightly (19.94~s) and deviated from the one with CORRECTION (14.98~s).
%		\item The completion time increased further with the ROLL angle being $45^\circ$ (22.70~s), $135^\circ$ (34.36~s), $225^\circ$ (19.94~s), and $315^\circ$ (27.34~s).
%		\item The completion time reached the maximum when the ROLL angle was $90^\circ$ (42.31~s) and $270^\circ$ (45.48~s).
%	\end{itemize}
\end{itemize}

\begin{figure}[tb]
	\centering
	\includegraphics[width=60mm]{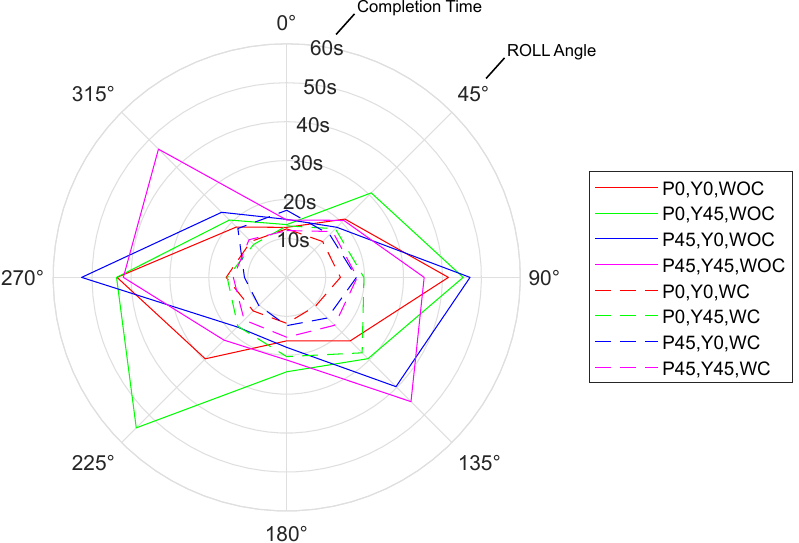}
	\caption{The three-way interaction effect of CORRECTION, ROLL, and PITCH-YAW. The red, green, blue, and magenta lines depict the average completion time of different ROLL angles with the PITCH-YAW angles being $0^\circ$ - $0^\circ$, $0^\circ$ - $45^\circ$, $45^\circ$ - $0^\circ$, and $45^\circ$ - $45^\circ$, respectively. The solid lines represent the results without CORRECTION, while the dashed lines correspond to the ones with CORRECTION. }
	\label{fig:simpolar}
\end{figure}

To examine the pattern of the three-way interaction effect more clearly, we draw the results in polar graphs in Fig. \ref{fig:simpolar}.
The red, green, blue, and magenta lines depict the average completion time of different ROLL angles with the PITCH-YAW angles being $0^\circ$ - $0^\circ$, $0^\circ$ - $45^\circ$, $45^\circ$ - $0^\circ$, and $45^\circ$ - $45^\circ$, respectively. 
The solid lines represent the results under WOC, while the dashed lines correspond to WC. 
It can be observed that:
\begin{itemize}
	\item When the CORRECTION was enabled, the average completion time was less than or approximately equal to 20~s in all the conditions of ROLL and PITCH-YAW angles, except the one with ROLL being $135^\circ$ and PITCH-YAW being $0^\circ$ - $45^\circ$ (27.40~s).
	\item The curves of conditions under WOC and the differences of them from conditions under WC are consistent when the ROLL angle was $0^\circ$, $90^\circ$, $180^\circ$, and $270^\circ$.
	\item The curves of conditions under WOC are inconsistent when the ROLL angle was $45^\circ$, $135^\circ$, $225^\circ$, and $315^\circ$. Some conditions of PITCH-YAW angles presented much higher average completion time than the others at some ROLL angles, while being similar to the others at other ROLL angles.
\end{itemize}

\subsection{Discussion}
There was a statistically significant difference between the completion time under WC and WOC (Fig. \ref{fig:simboxtotal}).
By enabling CORRECTION for the misalignment of the ROLL angle, the overall average completion time of the teleoperation task was reduced by 48.9\%.

Moreover, the CORRECTION method seemed to work for all the conditions of ROLL angles.
In Fig. \ref{fig:simpolartotal}, the average completion time under different ROLL angles was between 13.74 s and 17.50 s.
The CORRECTION was seldom affected by the PITCH-YAW angles, too.
In Fig. \ref{fig:simpolar}, the average completion time under different PITCH-YAW angles was less than or approximately equal to 20 s except one case with ROLL being $135^\circ$ and PITCH-YAW being $0^\circ$ - $45^\circ$ in which the average completion time was 27.40 s.
For all the other conditions, the average completion time was between 10.18 s and 20.35 s.

When the CORRECTION was not enabled, however, the completion of the teleoperation task was significantly affected by the misalignment between the camera frame and the teleoperation frame with respect to varying ROLL angle, which can be reflected by Fig. \ref{fig:simboxtotal} and Fig. \ref{fig:simpolartotal}.
Particularly, it is inferred from the results that for human operators, the misalignment in the orthogonal directions ($ROLL=90^\circ$ and $ROLL=270^\circ$) is more difficult to correct than that in the opposite direction ($ROLL=180^\circ$).
When the ROLL angle was $180^\circ$, which means the operator needs to move in the opposite direction as presented in the camera view to compensate the misalignment, the average completion time increased by 42.3\% compared to the time with ROLL angle being $0^\circ$.
When the ROLL angle was $90^\circ$ and $270^\circ$, which means the operator needs to move in the orthogonal dirrection to correct the misalignment, the increase of completion time was 201.9\% and 224.6\%, respectively.
When the ROLL angle was $45^\circ$, $135^\circ$, $225^\circ$, and $315^\circ$, which means there was a combination of misalignment in both the opposite direction and the orthogonal direction, the increase of average completion time was between the extreme conditions of pure opposite misalignment ($180^\circ$) and pure orthogonal misalignment ($90^\circ$ and $270^\circ$).

As the proposed CORRECTION method is only in the 2D plane of the camera view, it was also affected by the PITCH-YAW angles which resulted in 3D misalignment.
From Fig.~\ref{fig:simbox} and Fig.~\ref{fig:simpolar}, it can be seen that the average completion time differed significantly with respect to different PITCH-YAW angles when the ROLL angle was $45^\circ$, $135^\circ$, $225^\circ$, and $315^\circ$.
We infer that this is because, due to the coupling effect between the PITCH-YAW and ROLL angles, the proportion of the orthogonal component in the misalignment changed more significantly in some conditions comparing with others.
However, as there are currently only eight data points in each condition, further experiments are needed to analyze the effects of the PITCH-YAW angles. 

Overall, the proposed correction method can improve the teleportation by reducing the completion time in the simulated tasks under different conditions.

\section{Validation of the Correction Method on a Teleoperated Eye-in-Hand Robot}

\begin{figure*}[tb]
	\centering
	\includegraphics[width=120mm]{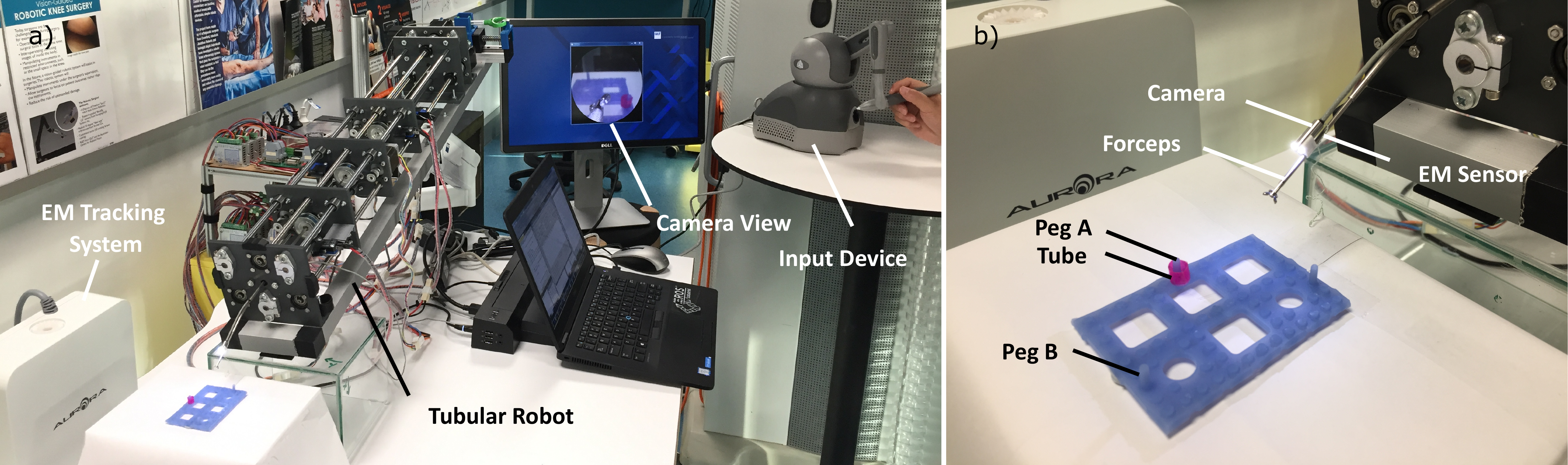}
	\caption{Experiment setup with a real \textit{eye-in-hand} robot. a) Full setup of the system. A tubular robot was teleoperated by an input device. The 3D motion of the tip of the robot was mapped by the 3D motion of the input device. A camera was rigidly attached to the tip of the robot and the view of the camera was displayed to the operator during teleoperation. b) Close-up of the task. The operator was required to teleoperate the robot from the initial position to move a pink tube from Peg A to Peg B. An electromagnetic (EM) sensor was mounted to the tip to track the trajectory of the motions.}
	\label{fig:experiment}
\end{figure*}

\begin{figure*}[tb]
	\centering
	\includegraphics[width=100mm]{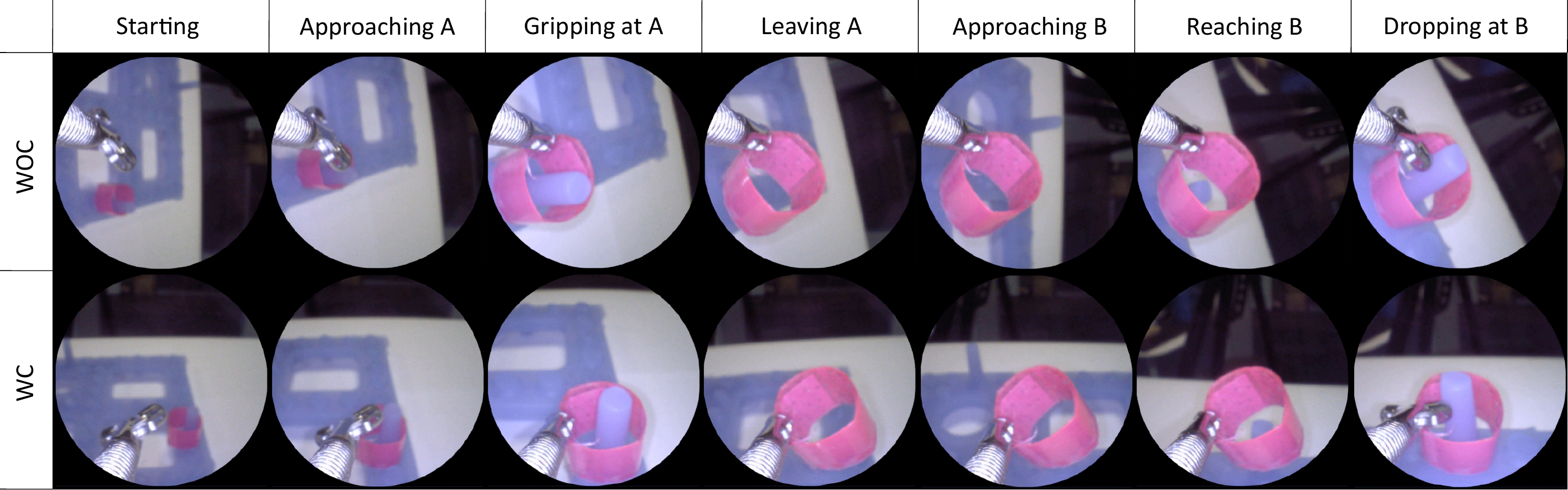}
	\caption{The tasks of the experiment. The robot should start from its initial position, grip the pink tube out of Peg A and move and place it on to Peg B. Each operator was required to teleoperate the robot to complete this task as fast as possible. The camera view was rotated around the axis of the lens for $90^\circ$ at the initial position to simulate the misalignment between the camera frame and the teleoperation frame. Two conditions were tested: one with the proposed correction method and one without. Each condition consisted of two pre-trials and five formal trials. The order was designed with a Latin Square method to balance the carryover effects like learning and fatigue.}
	\label{fig:process}
\end{figure*}

In this section, we further validate the proposed correction method by inviting participants to carry out a set of experiments using a real teleoperated \textit{eye-in-hand} robot.
An overall evaluation approach is taken because the real robot cannot be configured to have constant misalignment during the teleoperation like in the simulated environment discussed in Sec. \ref{sec:simulation}.

\subsection{Participants}
A total of 12 participants, who were a random subset of the 32 participants in the experiments described in the last section, volunteered for the test. The participants consisted of 10 males and 2 females, with ages ranging between 22 and 36 (M = 30.17, SD = 3.46). All the participants had no experience with teleoperation of a robot.
All the participants provided informed consent.
This protocol was approved by the Australian National Health and Medical Research Council (NHMRC) Registered Committee Number EC00171 under Approval Number 1800000891.

\subsection{Methods}
The participants were invited to perform on a real robotic system, as shown in Fig. \ref{fig:experiment}. 
The experimental setup included a tubular robot \cite{wu2016towards} that was teleoperated by an input device (The Touch, 3D Systems, Inc., USA). 
The robot had 5 degrees of freedom (DOFs) but only the 3 positional DOFs and the rotational DOF of the distal tube around its axis were telecontrolled. 
The distal tube was equipped with a pair of forceps for gripping objects. 
A micro camera was rigidly attached to the distal tube and the images captured by the camera were displayed to the operator. 
An electromagnetic (EM) sensor was also rigidly mounted to the distal tube and its position was tracked by an EM tracking system (Aurora, Northern Digital Inc., Canada) at a frequency of 20 Hz.

The tasks of the experiment are illustrated in Fig. \ref{fig:process}. 
The robot was required to start from its initial position, grip the pink tube out of Peg A and move and place it on to Peg B. 
Each operator was asked to teleoperate the robot to complete this task as fast as possible. 
The camera view was manually rotated around the axis of the lens for $90^\circ$ at the initial position to simulate the misalignment between the camera frame and the teleoperation frame.
This angle was chosen because it was one of the worst scenarios (most difficult to compensate by operators) based on simulation results in Section III.
Note, however, that the misalignment may continuously change during each experiment trial as the innermost tube may be rotated through users' teleoperation.
Two conditions were tested: one with the proposed correction method and one without. Each condition consisted of two pre-trials (not counted) and five formal trials (counted). 
The order was designed with a Latin Square method to balance the carryover effects like learning and fatigue \cite{yoshimura1996historical}.

After completing all the tasks, each operator was immediately asked to fill in a survey in which the difficulty of the task, the unsteadiness of the motion, and the mental stress experienced during the teleoperation, were rated for both conditions.
The original survey form can be found in the attachment of this article.

\subsection{Results}
\begin{figure}[tb]
	\centering
	\includegraphics[width=60mm]{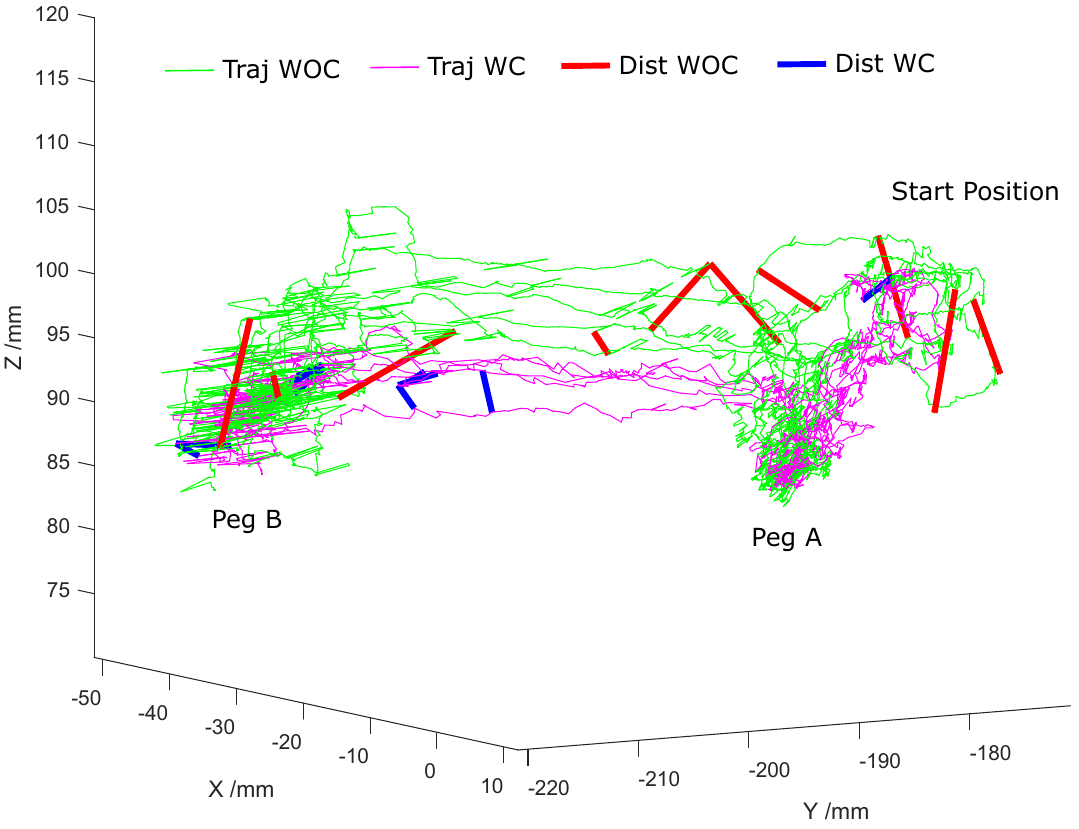}
	\caption{An example of the trajectories in two experimental conditions. Traj WOC, Traj WC, Dist WOC, and Dist WC stand for trajectory without correction, trajectory with correction, Hausdorff distance withouth correction, and Hausdorff distance with correction, respectively.}
	\label{fig:trajectory}
\end{figure}

Three indices, namely the completion time, the trajectory length, and the Hausdorff distance of the trajectories \cite{hausdorff}, were used to evaluate the performance of the tasks under different conditions.
The completion time was counted from the moment the robot started moving until the time the tube was successfully placed onto Peg B.
The trajectory length was calculated by accumulating the distances between adjacent points sampled at 20 Hz.
The Hausdorff distance, defined by
\begin{equation}
	d_H(X,Y) = \max\left\lbrace \sup_{x \in X}\inf_{y \in Y} d(x,y), \sup_{y \in Y}\inf_{x \in X} d(x,y)\right\rbrace, 
\end{equation}
was used to describe the consistency of the five trajectories of the trails in each condition.
As each trajectory was compared with the other four for the Hausdorff distance calculation, there were ten Hausdorff distances in total for each condition.
An example of the trajectories and the Hausdorff distance can be found in Fig. \ref{fig:trajectory}.

\begin{figure}[tb]
	\centering
	\includegraphics[width=60mm]{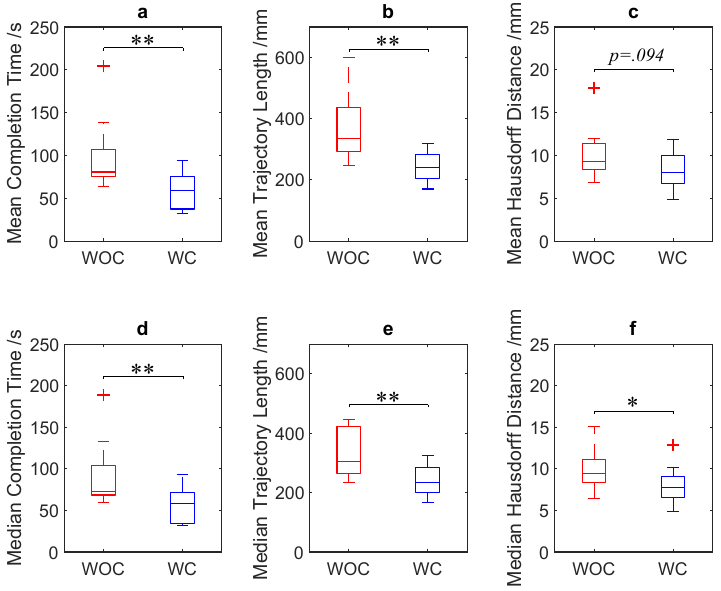}
	\caption{Paired-samples \textit{t}-test results of the Mean and Median values of the Completion Time, Trajectory Length, and Hausdorff Distance under WC and WOC. Each condition consists of 12 data points. *$p<0.05$, **$p<0.01$.}
	\label{fig:expbox}
\end{figure}

Paired-samples \textit{t}-tests were performed to the mean and median values of the completion time, trajectory length, and Hausdorff distance of the five trials in each condition conducted by the 12 participants.
The results are plotted in Fig. \ref{fig:expbox}.
There was a statistically significant difference between WC and WOC in five comparisons: 
\begin{itemize}
	\item  Mean Completion Time, $58.16 \pm 22.18$ s vs. $97.93 \pm 39.54$~s, $t(11)=-4.742$, $p=0.001$; 
	\item  Median Completion Time, $56.10 \pm 21.63$ s vs. $91.41 \pm 37.67$ s, $t(11)=-4.780$, $p=0.001$;
	\item  Mean Trajectory Length, $242.05 \pm 45.74$ mm vs. $368.14 \pm 102.99$ mm, $t(11)=-4.627$, $p=0.001$;
	\item  Median Trajectory Length, $238.57 \pm 50.63$ mm vs. $331.71 \pm 77.95$ mm, $t(11)=-4.737$, $p=0.001$;
	\item  Median Hausdorff Distance, $8.00 \pm 2.11$ mm vs. $9.92 \pm 2.34$ mm, $t(11)=-2.449$, $p=0.032$.
\end{itemize}

For the Mean Hausdorff Distance, the difference was not statistically significant, $8.35 \pm 2.18$ mm vs. $10.13 \pm 2.93$ mm, $t(11)=-1.831$, $p=0.094$.

\begin{figure}[tb]
	\centering
	\includegraphics[width=60mm]{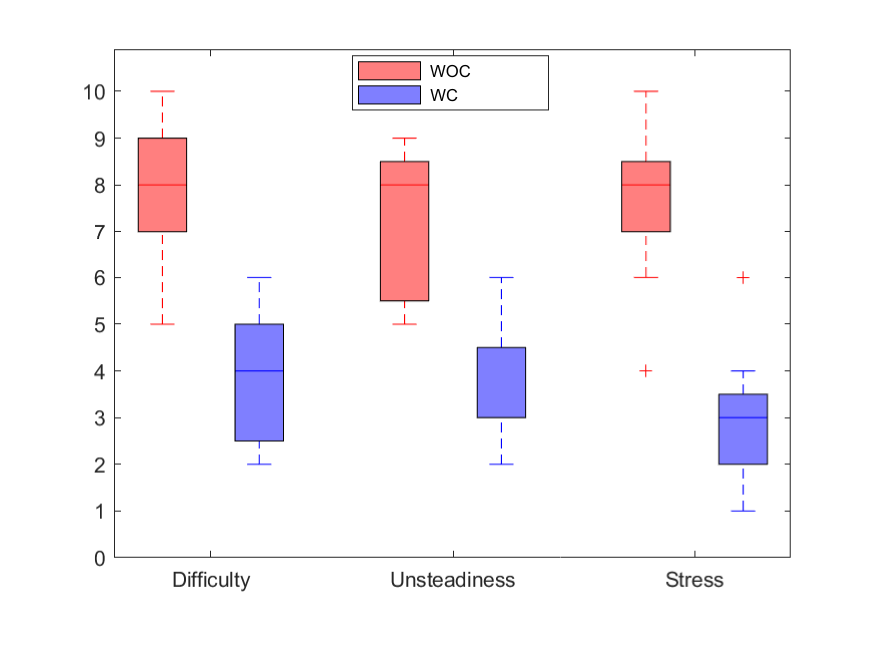}
	\caption{The results of the survey after completion of the tasks. The operators were asked to rate the difficulty of the task, the unsteadiness of the motion, and the mental stress experienced during the teleoperation for both conditions on a 1 - 10 scale, where 1 indicates the lowest difficulty/unsteadiness/stress and 10 indicates the highest difficulty/unsteadiness/stress. ***$p<0.001$.}
	\label{fig:expsurvey}
\end{figure}

The results of the survey after the completion of the tasks were displayed in Fig. \ref{fig:expsurvey}.
Paired-sample \textit{t}-tests were performed to compare the difficulty of the task, the unsteadiness of the motion, and the mental stress experienced during the teleoperation for both conditions.
There were significant differences in all the three aspects compared between WC and WOC:
\begin{itemize}
	\item Difficulty, $3.83 \pm 1.47$ vs. $7.75 \pm 1.42$, $t(11)=-7.035$, $p<0.001$;
	\item Unsteadiness, $3.67 \pm 1.37$ vs. $7.25 \pm 1.60$, $t(11)=-5.77$, $p<0.001$;
	\item Stress, $3.00 \pm 1.28$ vs. $7.67 \pm 1.56$, $t(11)=-9.38$, $p<0.001$.
\end{itemize}

\subsection{Discussion}
From the results in Fig. \ref{fig:expbox}, it is clear that the proposed correction method improved the teleportation in terms of completion time and trajectory length when there was an initial misalignment of $90^\circ$ in the roll angle.
With the correction enabled, the mean completion time was reduced by 40.6\% (38.6\% for median values) and the mean trajectory length was shorten by 34.3\% (28.1\% for median values).
This indicates that the correction of the misalignment helped the operators find a more straightforward path to complete the transfer task with a shorter time.

As for the consistency of the trajectories, it is interesting to see there was a statistically significant difference between the median Hausdorff Distances of the trajectories under WC and WOC, while no significant difference in the mean values.
The disparity between the mean and median values could be caused by extreme values since the median is more robust against extremely high or low values.
The result shows that the improvement of trajectory consistency under WC in the experiments was slight and around the threshold of statistical significance.

However, the operators experienced significantly differently with the two conditions.
The subjective evaluation results plotted in Fig. \ref{fig:expsurvey} show clear distinction in terms of difficulty of the task, unsteadiness of the motion, and mental stress experienced during the teleoperation.
The operation with CORRECTION enabled saw an improvement in the three aspects by 50.5\%, 49.4\%, and 60.9\%, respectively.
These results further validate the effectiveness of the proposed correction for the misalignment of the camera frame.

\section{Conclusion}
In this paper we investigated the effects of the camera frame misalignment in a teleoperated \textit{eye-in-hand} system.
Through a user study on a simulated system in which there was misalignment in both the ROLL angle and the PITCH-YAW angles, we found that the operator performance evinced a significant drop with the ROLL angle being $90^\circ$ and $270^\circ$.
The results suggested that misalignment between the input motion and the output view is much more difficult to compensate by the operators when it is in the orthogonal direction than when it is in the opposite direction.
It was also observed that there was a coupling between the misalignment in the ROLL angles and the PITCH-YAW angles.

A simple correction method in the view display was proposed and tested.
Through the same simulation study, it was validated that the correction method worked for all the conditions tested.

A comparison between with and without the proposed correction was further conducted on a real tubular robot.
The results showed that with the correction enabled, there was a significant improvement in the operation performance in terms of completion time (mean 40.6\%, median 38.6\%), trajectory length (mean 34.3\%, median 28.1\%), difficulty (50.5\%), unsteadiness (49.4\%), and mental stress (60.9\%). 
This suggests that the proposed correction method could improve the use of the compact \textit{eye-in-hand} tubular robot in a confined environment such as in endoscopic surgery.

The main limitations of this work include two aspects.
First, the coupling between the ROLL angle and the PITCH-YAW angles was not fully investigated due to limited samples and trials.
Second, the PITCH-YAW angles have been restricted to be selected from the binary set $\{0^\circ, 45^\circ\}$, which conformed with the capability and working conditions of the tubular robot system that motivated this study.
However, more configurations are needed to reveal the effects of the full spectrum of the PITCH-YAW angles.
Therefore, our future work will be to investigate the effect resulted from the PITCH-YAW angles with more user studies.
A combination of correction in both view display and motion mapping can also be considered with this more complex problem.
Moreover, the correction has been limited to the one DOF of rotation around the camera axis.
Although the results have shown the effectiveness of this simple method, a correction involving more DOFs can be investigated.

%A conclusion section is not required. Although a conclusion may review the main points of the paper, do not replicate the abstract as the conclusion. A conclusion might elaborate on the importance of the work or suggest applications and extensions. 

%                                  % on the last page of the document manually. It shortens
%                                  % the textheight of the last page by a suitable amount.
%                                  % This command does not take effect until the next page
%                                  % so it should come on the page before the last. Make
%                                  % sure that you do not shorten the textheight too much.

%%%%%%%%%%%%%%%%%%%%%%%%%%%%%%%%%%%%%%%%%%%%%%%%%%%%%%%%%%%%%%%%%%%%%%%%%%%%%%%%

%%%%%%%%%%%%%%%%%%%%%%%%%%%%%%%%%%%%%%%%%%%%%%%%%%%%%%%%%%%%%%%%%%%%%%%%%%%%%%%%

%%%%%%%%%%%%%%%%%%%%%%%%%%%%%%%%%%%%%%%%%%%%%%%%%%%%%%%%%%%%%%%%%%%%%%%%%%%%%%%%
%\section*{APPENDIX}

\section*{ACKNOWLEDGMENT}

The authors thank Mr. Lucian Quach and Ms. Yuting Sun for their help in the recruitment of participants. L. Wu thanks Dr. Si Wang for her advice on the experiment design and statistics.

%%%%%%%%%%%%%%%%%%%%%%%%%%%%%%%%%%%%%%%%%%%%%%%%%%%%%%%%%%%%%%%%%%%%%%%%%%%%%%%%

%References are important to the reader; therefore, each citation must be complete and correct. If at all possible, references should be commonly available publications.

\bibliographystyle{IEEEtran}
\bibliography{LW}

%%%%%%%%%%%%%%%%%%%%%%%%%%%%%%%%%%%%%%%%%%%%%%%%%%%%%%%%%%%%%%%%%%%%%%%%%%%%%%%%

\begin{IEEEbiography}[{\includegraphics[width=1in,height=1.25in,clip,keepaspectratio]{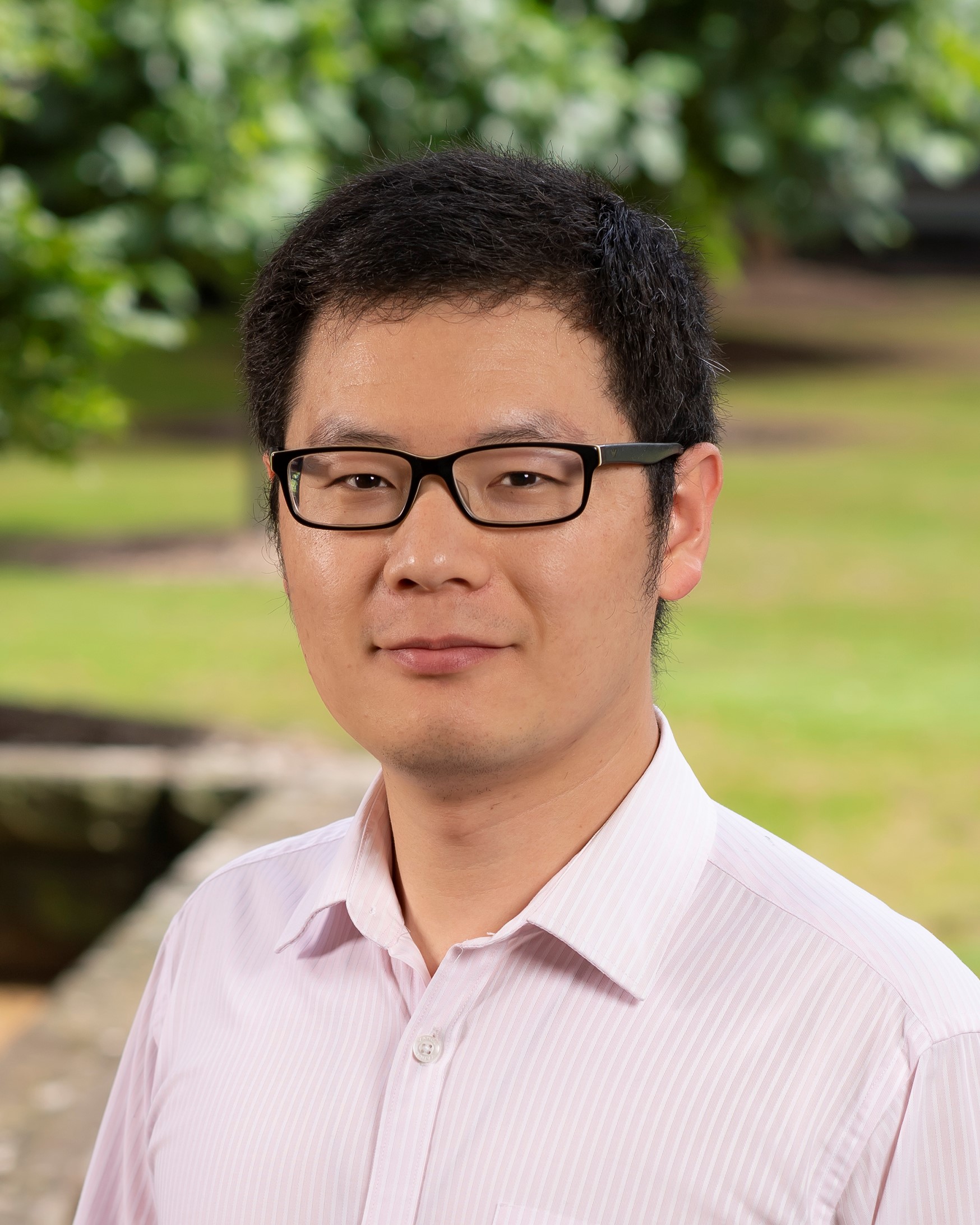}}]{Liao Wu} (Member, IEEE) received the B.S. and
Ph.D. degrees in mechanical engineering from Tsinghua University, Beijing, China, in 2008 and 2013, respectively.

He is currently a Senior Lecturer with the School of Mechanical and Manufacturing Engineering, University of New South Wales, Sydney, NSW, Australia. From
2014 to 2015, he was a Research Fellow with the National University of Singapore, Singapore. He then worked as a Vice-Chancellor’s Research Fellow with
the Queensland University of Technology, Brisbane, QLD, Australia, from 2016 to 2018. Between 2016 and 2020, he was affiliated with the Australian Centre for Robotic Vision (formerly ARC Centre of Excellence), Brisbane, QLD, Australia. He has worked on applying Lie groups theory to robotics, kinematic modeling and calibration, etc. His current research focuses on medical robotics, including flexible robots and intelligent perception for minimally invasive surgery.
\end{IEEEbiography}

\begin{IEEEbiography}[{\includegraphics[width=1in,height=1.25in,clip,keepaspectratio]{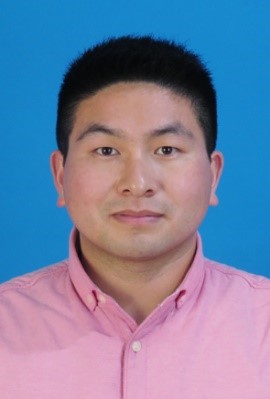}}]{Fangwen Yu}%
is currently a postdoctoral research fellow in the Center for Brain Inspired Computing Research and the Department of Precision Instrument at Tsinghua University, Beijing, China. He received his PhD from the China University of Geosciences, Wuhan, China. 

His research interests include brain-inspired 3D navigation, neuromorphic 3D SLAM, event-based 3D spatial sensing, etc. He is a member of IEEE, the IEEE Robotics and Automation Society (RAS) and the Royal Institute of Navigation (RIN). 
\end{IEEEbiography}

\begin{IEEEbiography}[{\includegraphics[width=1in,height=1.25in,clip,keepaspectratio]{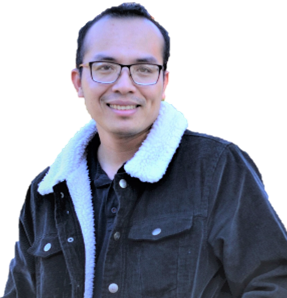}}]{Thanh Nho DO} 
	is currently a Scientia Senior Lecturer at the Graduate School of Biomedical Engineering, UNSW, Sydney, Australia. He is the founder of UNSW Medical Robotics Lab. In 2015, he was awarded his PhD degree in Surgical Robotics from the School of Mechanical \& Aerospace Engineering (MAE), Nanyang Technological University (NTU), Singapore. He was a Postdoctoral Scholar at California NanoSystems Institute, University of California Santa Barbara, USA. He also worked as a Research Fellow at the Robotic Research Center, School of MAE, NTU, Singapore. His research interests include soft robotics, wearable haptics, control, surgical robotics, and mechatronics in medicine. 
\end{IEEEbiography}

\begin{IEEEbiography}[{\includegraphics[width=1in,height=1.25in,clip,keepaspectratio]{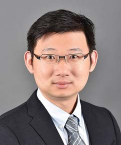}}]{Jiaole Wang}
received the B.E. degree in mechanical engineering from Beijing Information Science and Technology University, Beijing, China, in 2007, the M.E. degree from the Department of Human and Artificial Intelligent Systems, University of Fukui, Fukui, Japan, in 2010, and the Ph.D. degree from the Department of Electronic Engineering, The Chinese University of Hong Kong (CUHK), Hong Kong, in 2016. 
	
He was a Research Fellow with the Pediatric Cardiac Bioengineering Laboratory, Department of Cardiovascular Surgery, Boston Children’s Hospital and Harvard Medical School, Boston, MA, USA. He is currently an Associate Professor with the School of Mechanical Engineering and Automation, Harbin Institute of Technology, Shenzhen, China. His main research interests include medical and surgical robotics, image-guided surgery, human-robot interaction, and magnetic tracking and actuation for biomedical applications.
\end{IEEEbiography}

\end{document}